\documentclass[conference,10pt]{IEEEtran}
\usepackage{epsfig,rotating,setspace,latexsym,amsmath,epsf,amssymb,amsfonts,bm,theorem,subfigure,epstopdf}
\usepackage{cite,authblk}
\usepackage{bbm}
\usepackage{color}
\usepackage{mathtools}
\usepackage{algorithm}
\usepackage[noend]{algpseudocode}
\usepackage{verbatim}

\algrenewcommand\algorithmicforall{\textbf{foreach}}
\algrenewcommand\algorithmicindent{.8em}

\IEEEoverridecommandlockouts
\allowdisplaybreaks

\begin{document}

\title{Hierarchical Over-the-Air FedGradNorm}

\author{Cemil Vahapoglu \qquad Matin Mortaheb  \qquad Sennur Ulukus\\
\normalsize Department of Electrical and Computer Engineering\\
\normalsize University of Maryland, College Park, MD 20742\\
\normalsize  \emph{cemilnv@umd.edu} \qquad \emph{mortaheb@umd.edu}  \qquad \emph{ulukus@umd.edu}}

\maketitle

\begin{abstract}
Multi-task learning (MTL) is a learning paradigm to learn multiple related tasks simultaneously with a single shared network where each task has a distinct personalized header network for fine-tuning. MTL can be integrated into a federated learning (FL) setting if tasks are distributed across clients and clients have a single shared network, leading to personalized federated learning (PFL). To cope with statistical heterogeneity in the federated setting across clients which can significantly degrade the learning performance, we use a distributed dynamic weighting approach. To perform the communication between the remote parameter server (PS) and the clients efficiently over the noisy channel in a power and bandwidth-limited regime, we utilize over-the-air (OTA) aggregation and hierarchical federated learning (HFL). Thus, we propose hierarchical over-the-air (HOTA) PFL with a dynamic weighting strategy which we call \emph{HOTA-FedGradNorm}. Our algorithm considers the channel conditions during the dynamic weight selection process. We conduct experiments on a wireless communication system dataset (RadComDynamic). The experimental results demonstrate that the training speed with \emph{HOTA-FedGradNorm} is faster compared to the algorithms with a naive static equal weighting strategy. In addition, \emph{HOTA-FedGradNorm} provides robustness against the negative channel effects by compensating for the channel conditions during the dynamic weight selection process.
\end{abstract}
 
\section{Introduction}

Federated learning (FL) is a distributed learning framework where many clients train a shared model under the orchestration of a centralized server while keeping the training data decentralized and private. In addition to FL, multi-task learning (MTL) is a learning paradigm that aims to learn multiple related tasks simultaneously by learning a shared representation for all tasks \cite{MTLintro, Zhang2017ASO}. Integration of MTL in the FL framework leads to personalized federated learning (PFL), in which clients have different tasks, and the clients train the common shared network under the orchestration of the centralized server while each client further trains a small client-specific network for its own specific task, referred to as personalization. 
The clients in an FL setting can have non-iid training data. Further, task complexities across clients can be different. Both the non-iid nature of the training data across clients and different task complexities cause statistical heterogeneity, leading to overall system performance degradation. Distributed dynamic weighting strategy, called \emph{FedGradNorm} is previously proposed to handle statistical heterogeneity in PFL by balancing the learning speeds across different tasks \cite{FedGradNorm}. \emph{FedGradNorm} utilizes the clients’ gradients on the server-side before aggregation to be able to do this.

\begin{figure}[t]
 \centerline{\includegraphics[width=1\linewidth]{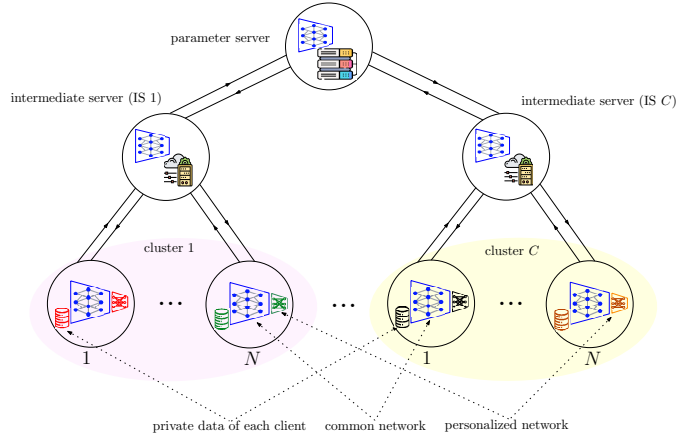}}
  \caption{Hierarchical personalized federated learning framework with a common network (shown in blue) and small personalized headers (shown in red, green, black, orange).}
  \label{HOTA_FLgrad}
\end{figure}

In \cite{FedGradNorm}, the characteristics of the communication channel are not considered, although it is known that the clients can be distributed by geographic location in FL \cite{Hsieh2017GaiaGM, YANG202233}. In certain applications, the parameter server (PS) can be far away from the clients, making the communication between the PS and the clients noisy and vulnerable to channel effects. The communication over a shared wireless channel needs to be done over a power- and bandwidth-limited setting, bringing communication cost concerns into FL. To address these issues, over-the-air (OTA) aggregation \cite{Amiri2019OvertheAirML, Amiri2019MachineLA} has become a prominent approach as an efficient strategy for supporting clients on the same bandwidth. In addition, hierarchical federated learning (HFL) framework is introduced by creating clusters of clients around intermediate servers (IS) which communicate with the PS instead of direct communication of clients with the PS. The studies about HFL focus on different aspects such as latency and power analysis \cite{Abad2020HierarchicalFL,Liu2020ClientEdgeCloudHF}, resource allocation \cite{Luo2020HFELJE, Wang2020LocalAH}. These works consider the HFL with error-free channels. \cite{Aygn2021HierarchicalOF} investigates HFL with OTA by taking into account the practical wireless channel models. \cite{OTAPerFed} investigates PFL with OTA across clusters in which the clients inside the cluster are assumed to have the same data distribution. In contrast, statistical heterogeneity may exist since clients may be responsible for different tasks, especially when clients are grouped based on their geo-locations. 

In our work, we introduce hierarchical over-the-air personalized federated learning with \emph{FedGradNorm}, which we call \emph{HOTA-FedGradNorm}, by adopting \emph{FedGradNorm} into HFL by utilizing OTA aggregation for the bandwidth-limited wireless fading multiple access channel (MAC) when the transmission power constraints are imposed on clients. \emph{FedGradNorm} \cite{FedGradNorm} is modified by taking into account the channel conditions, thereby imposing implicit constraints on the dynamic weighting coefficients. The experiments are conducted to demonstrate how \emph{HOTA-FedGradNorm} deals with the effect of the channel condition via the dynamic weight selection process. We conduct experiments on a wireless communication system dataset called RadComDynamic. Our experimental results demonstrate that the training speed with \emph{HOTA-FedGradNorm} is faster compared to algorithms with a naive static equal weighting strategy. In addition, we observe that \emph{HOTA-FedGradNorm} provides robustness against the negative channel effects by compensating for the channel conditions during the dynamic weight selection process. Finally, extended results for this work and \cite{FedGradNorm} are presented in \cite{mortaheb2022personalized} with theoretical analysis.

\section{System Model and Problem Formulation}

The generic form of HFL problem with $C$ clusters each containing an intermediate server (IS) and $N$ clients as depicted in Fig.~\ref{HOTA_FLgrad} is,
\begin{align}
        \min_{\omega} \left\{ F(\omega) \triangleq \frac{1}{CN} \sum_{l=1}^C \sum_{i=1}^N p^{(l,i)}F^{(l,i)}(\omega)\right\}
\end{align}
where $p^{(l,i)}$ is the loss weight for client $i$ in cluster $l$ such that $\sum_{i=1}^N p^{(l,i)}= N$, $\forall l \in [C]$, and $F^{(l,i)}(\cdot)$ is the local loss function for client $i$ in cluster $l$.

We consider a PFL setting of $N$ clients within each cluster, in which client $i$ of cluster $l$ has its own local dataset $D_{l,i} = \{(\mathbf{x}^{(l,i)}_j,y^{(l,i)}_j) \}_{j=1}^{n_{l,i}}$ where $n_{l,i}$ is the size of the local dataset. Within cluster $l$, $T_{l,i}$ denotes the task of client $i$, $i \in [N]$, and $l \in [C]$. $T_{l,i}$ is assigned from the task set $\mathcal{T} = \{T_1 , T_2, \ldots, T_N \}$ such that $T_{l,i} \neq T_{l,i'}$, for $i \neq i' $ and any $l \in [C]$. Real-life scenarios might involve the same or very similar tasks for clients in a cluster. We assume that tasks are different due to the lack of prior information about them.

Additionally, clients in a cluster are connected to their corresponding intermediate server via local area network (LAN), which are assumed to be error-free and to have high data transmission rate. The ISs are connected to the PS via bandwidth-limited fading MAC for sending the local gradient aggregations in clusters to the PS. The broadcast from the PS to the ISs is considered to be error-free.

The system model in Fig.~\ref{HOTA_FLgrad} is composed of a global representation network $q_\omega :\mathbb{R}^{d} \rightarrow \mathbb{R}^{d'}$, which is a function parameterized by $\omega \in \mathcal{W}$, that maps data points into a lower space of size $d'$. All clients in each cluster share the same global representation network, which is synchronized through global aggregation. The client-specific head $q_{h^{(l,i)}} : \mathbb{R}^{d'} \rightarrow \mathcal{Y}$ is a function parameterized by $h^{(l,i)} \in \mathcal{H}$ for all clients $i \in [N]$ of every cluster $l \in [C]$, mapping a low dimensional representation space to the label space $\mathcal{Y}$. The local model for client $i$ of cluster $l$ is the composition of the client's global representation model $q_\omega$ and  personalized model $q_{h^{(l,i)}}$,  shown as $q_{l,i} (\cdot) = (q_{h^{(l,i)}} \circ q_\omega)(\cdot)$. In addition, the local loss for the $i$th client of cluster $l$ is shown as $F^{(l,i)}(h^{(l,i)},\omega) = F^{(l,i)}(q_{l,i} (\cdot)) = F^{(l,i)}((q_{h^{(l,i)}} \circ q_\omega)(\cdot))$.

Using alternating minimization, the PS and the clients learn the global representation $\omega$ together, while only client $i$ learns the the client-specific head $h^{(l,i)}$ in cluster $l$, $i \in [N]$ and $l \in [C]$. Specifically, client $i$ of cluster $l$ performs $\tau_h$ local updates to optimize $h^{(l,i)}$ when global network parameters at client $i$ of cluster $l$, i.e., $ \omega^{(l,i)}$ are frozen. Then, $\tau_\omega$ local updates are performed to optimize $\omega^{(l,i)}$ while the parameters corresponding to the client-specific head are frozen. Thereafter, the $l$th IS aggregates $\{ \omega^{(l,i)} \}_{i=1}^N$ which are sent via LAN, for any $l \in [C]$. The ISs send cluster aggregations to the PS to perform the global aggregation over the wireless fading MAC. The global aggregation is performed over the air by the additive nature of wireless MAC. Considering the overall system model scheme, the optimization problem is
\begin{align}
    \min_{\omega \in \mathcal{W}} \frac{1}{CN} \sum_{l=1}^C \sum_{i=1}^N p^{(l,i)} \min_{h^{(l,i)} \in \mathcal{H}} F^{(l,i)}(h^{(l,i)},\omega)
\end{align}

\section{Algorithm Description}
The algorithm consists of two stages. In the first stage, a distributed dynamic weighting strategy is applied to balance learning speed of tasks across clients within each cluster. Dynamic weighting strategy is combined with power allocation scheme to satisfy the total average transmit power constraint and to be robust against the channel effects of bandwidth-limited fading MAC between the ISs and the PS. In the second stage of the algorithm, the global aggregation is performed over the air by utilizing the additive nature of wireless MAC. Then, the aggregated gradient is estimated on the PS to update the global representation network.

\subsection{Local Aggregation via Dynamic Weighting}
During the first stage of the algorithm, each client within a cluster sends its gradient for the global model $q_\omega$ to its corresponding IS via LAN, where the channels between each client and the corresponding IS are assumed to be error-free inside a cluster. Then, the corresponding IS performs a modified version of \emph{FedGradNorm} \cite{FedGradNorm} as a distributed dynamic weighting strategy  based on the client's gradients by taking taking the power allocation scheme into account to balance the learning speed across different tasks. 

Specifically, the IS of cluster $l$ computes the loss weight $p_k^{(l,i)}$ for each client $i \in [N]$ in cluster $l$ via \emph{FedGradNorm} algorithm to eventually obtain the local weighted aggregation $\sum_{i=1}^N p_k^{(l,i)}g_k^{(l,i)}$ at iteration $k$, where $g_k^{(l,i)}$ is the local gradient update of client $i$ in cluster $l$ for iteration $k$. Additionally, the power allocation vector $\beta_k^{(l,i)}$ constructed by the IS of cluster $l$ for each client $i$ in the cluster is designed as:
\begin{align}\label{power_vector}
    \beta_k^{(l,i)}(j) = \begin{cases}
        \frac{p_k^{(l,i)}}{H_k^{l}(j)}, & \text{if} \; |H_k^{(l)}(j)|^2 \geq H_k^{\textrm{th}},\\
        0, & \; \text{otherwise}
    \end{cases}
\end{align}
where $\beta_k^{(l,i)}(j)$ is the $j$th entry of the power allocation vector $\beta_k^{(l,i)} \in \mathbb{R}^{|\omega|}$, and $H_k^{(l)}(j)$ is the $j$th entry of the channel gain vector $H_k^{(l)} \in \mathbb{R}^{|\omega|}$, which represents the effect of the wireless fading channel between the IS of cluster $l$ and the PS. $H_k^{(l)}(j)$ is assumed to be independent and identically distributed (iid) according to $\mathcal{N}(0,\sigma_l^2)$.  The threshold $H_k^{\textrm{th}}$ is set to satisfy the average transmit power constraint given as follows,
\begin{align}\label{power constraint}
    \frac{1}{K} \sum_{k=1}^K \mathbb{E}[\|x_{k}^{(l)}\|^2] \leq \Bar{P}
\end{align}
where $x_k^{(l)} = \sum_{i=1}^N x_k^{(l,i)}$ and $x_k^{(l,i)} = \beta_k^{(l,i)} \circ g_k^{(l,i)}$, $i \in [N], l \in [C]$, $\circ$ represents the element-wise multiplication. The expectation is taken over the randomness of the channel gains. 

From the power allocation scheme in (\ref{power_vector}), each cluster transmits only the scaled entries of its weighted gradient for which the corresponding channel conditions are sufficiently good. It provides implicit gradient sparsification by saving transmission power. Therefore, we can also modify \emph{FedGradNorm} algorithm by sparsifying the auxiliary loss function $F_{\textrm{grad}}$ in \cite{FedGradNorm} before minimizing it to have $\{ p_k^{(l,i)}\}_{i=1}^N, \forall l \in [C]$. Sparsified $F_{\textrm{grad}}$ for the cluster $l$ is the following,
\begin{align}\label{F_grad_sparsified}
    &F_{\textrm{grad}}^{(l)} \left(k ; \{p^{(l,i)}_k\}_{i=1}^N \right)=\sum_{i=1}^N F^{(l,i)}_{\textrm{grad}} \left(k ; p^{(l,i)}_k\right) \\
    &=\sum_{i=1}^N  \left \| p^{(l,i)}_k  \left\| \textrm{M}^{(l)}_k \circ \nabla_{\tilde{\omega}^{(l,i)}_k} F^{(l,i)}_k \right\| - \bar{G}^{(l)}_{\tilde{\omega}} (k) \times [r^{(l,i)}_k]^\gamma \right \|
\end{align}
where $M^{(l)} \in \{0,1\}^{|\omega|}$ is a mask matrix designed for the sparsification of cluster $l$ as follows:
\begin{align}
    \textrm{M}_k^{(l)}(j) = \begin{cases}
    1, & \text{if} \; |H_k^{(l)}(j)|^2 \geq H_k^{\textrm{th}},\\
    0, & \; \text{otherwise}
\end{cases}
\end{align}
Here, $\tilde{\omega}^{(l,i)}_k$ is the last layer of the shared network at client $i$ of cluster $l$ at iteration $k$. $\bar{G}^{(l)}_{\tilde{\omega}} (k)$ is the average sparsified gradient norm across all clients (tasks) in cluster $l$ at iteration $k$. $r^{(l,i)}_k =\frac{\tilde{F}^{(l,i)}_k}{\mathbb{E}_{j \sim \textrm{task}}[\tilde{F}^{(l,j)}_k]}$ is the relative inverse training rate of task $i$ in cluster $l$ at iteration $k$, and $\gamma$ represents the strength of the restoring force which pulls tasks back to a common training rate, which can also be thought of as a metric of task asymmetry across different tasks.  

Gradient sparsification used during the calculation of $F_{\textrm{grad}}$ acts as an implicit constraint on $F_{\textrm{grad}}$ minimization problem by considering the channel conditions. Consequently, it ensures that the learning speed of tasks is invariant to the dynamic channel conditions with an appropriate selection process of loss weights. In other words, the implicit constraint of the channel condition preserves the fairness of the learning speed among the clients, as shown in the experimental results.

\subsection{Over-the-Air Aggregation}
The second stage of the algorithm involves the process of global aggregation over the wireless fading MAC. The PS obtains a noisy estimate of the aggregated gradient over the wireless fading channel while updating the model parameters. Due to the additive nature of the wireless MAC, the summation of the signals transmitted by clusters arrives at the PS. The $j$th entry of the received signal at iteration $k$, $y_k \in \mathbb{R}^{|\omega|}$ is
\begin{align}\label{recevied_signal}
    y_k(j) &= \sum_{l \in \mathcal{M}_k(j)} H_k^{(l)}(j)x_k^{(l)}(j)+ z_k(j)
\end{align}
where $z_k(j)$ is the $j$th entry of the Gaussian noise vector $z_k$ and is iid according to $\mathcal{N}(0,1)$. $\mathcal{M}_k(j) = \{c \in [C] : |H_k^{(l)}(j)|^2 >H_k^{th} \}$ represents the set of clusters contributing to the $j$th entry of the received signal at the $k$th iteration. $\mathcal{M}_k(j)$ is known by the PS, for $j \in [|\omega|]$ since the PS has the perfect channel state information (CSI).

By considering (\ref{power_vector}) and the definition of $x_k^{(l)}$ in terms of the power allocation vector, we have
\begin{align} \label{received_signl_final}
    y_k(j) &= \sum_{l \in \mathcal{M}_k(j)} \sum_{i=1}^N p_k^{(l,i)}g_k^{(l,i)}(j) + z_k(j)
\end{align}
where $g_k^{(l,i)}(j)$ is the $j$th entry of $g_k^{(l,i)}$. The noisy aggregated gradient estimate is
\begin{align}\label{noisy estimate}
    \hat{g}_k(j) = \frac{y_k(j)}{|\mathcal{M}_k(j)|N}, \quad j \in [|\omega|]
\end{align}
Then, the estimated gradient vector is used to update the model parameters as $\omega_{k+1} = \omega_k - \beta \hat{g}_k$. The overall algorithm is shown in Algorithm~\ref{alg:hota-fedgradnorm}.

\begin{algorithm}[]
    \caption{HOTA-FedGradNorm}
    \label{alg:hota-fedgradnorm}
\begin{algorithmic}[1]
    \State Initialize $\omega_0$, $\{p_0^{(1,i)}\}_{l=1,i=1}^{C,N}$, $\{h_0^{(1,i)}\}_{l=1,i=1}^{C,N}$
    \For {$k$=0 {\bfseries to} $K$} 
        \State The PS broadcasts the current global shared network parameters $\omega_{k}$ to the ISs.
        \For {Each cluster $l \in [C]$} 
            \State $\omega_{k}^{(l)} \leftarrow \omega_{k}$ .
            \State The IS $l$ broadcasts $\omega_k^{(l)}$ to clients within cluster.
            \For{Each client $i \in [N]$} 
                \State Initialize global shared network parameters for local updates by $\omega_{k,0}^{(l,i)} \leftarrow \omega^{(l)}_{k}$
                \State Initialize $F_{k}^{(l,i)}=0$, and $g_{k}^{(l,i)}=0$
                
                \For{$j=1,\ldots,\tau_h$} 
                    \State $h_{k,j}^{(l,i)}$ =$\textrm{Update}(F^{(l,i)}(h_{k,j-1}^{(l,i)},\omega_{k,0}^{(l,i)}),h_{k,j-1}^{(l,i)})$
                \EndFor 
                \For{$j=1,\ldots,\tau_\omega$} 
                    \State $\omega_{k,j}^{(l,i)} \leftarrow \omega_{k,j-1}^{(l,i)} - \beta g_{k,j}^{(l,i)}$
                    \State $F_{k}^{(l,i)}$ += $\frac{1}{\tau_\omega}$ $F^{(l,i)}(h_{k,\tau_h}^{(l,i)},\omega_{k,j}^{(l,i)})$
                \EndFor 
                
                \State Client sends $g_{k}^{(l,i)} = \frac{1}{\tau_\omega} \sum_{j=1}^{\tau_\omega} g_{k,j}^{(l,i)}$, and $\tilde{F}_{k}^{(l,i)} = \frac{F_{k}^{(l,i)}}{F_{0}^{(l,i)}}$ to the IS $l$ for dynamic weighting.
            \EndFor 
            
            \State The IS $l$ performs the followings:
                \begin{itemize}
                    \item $\{p_{k}^{(l,i)}\}_{i=1}^N$=FGN\_server($\{g_{k}^{(l,i)} \}_{i=1}^N$,$\{\tilde{F}_{k}^{(l,i)} \}_{i=1}^N$,$p_{k-1}^{(l,i)}$)
                    \item The IS $l$ constructs the power allocation vector $\beta_{k}^{(l,i)}$ for each clients in cluster $l$ as given in eq. (\ref{power_vector}) 
                    \item aggregates the gradients of clients in cluster $l$ for the global shared network by combining with power allocation scheme as $x_k^{(l)} = \sum_{i=1}^N \beta_k^{(l,i)} \circ g_k^{(l,i)}$. 
                \end{itemize}
        \EndFor 
        \State The gradients are aggregated over the wireless fading channel as given in eq. (\ref{recevied_signal}).
        \State The estimated gradient aggregation $\Hat{g}_k$ is obtained by the PS as given in eq. (\ref{noisy estimate}).
        \State The PS updates the global shared network by $\omega_{k+1} \leftarrow \omega_k - \beta \Hat{g}_k$.
    \EndFor 
\end{algorithmic}
\end{algorithm}

$\textrm{Update}(f,h)$ in Algorithm~\ref{alg:hota-fedgradnorm} represents the generic notation for the update of the variable $h$ by using the gradient of $f$ function with respect to the variable $h$. $\omega_{k,j}^{(l,i)}$, $h_{k,j}^{(l,i)}$, and $g_{k,j}^{(l,i)}$ denote the global shared network parameters, the client-specific network  parameters and the gradient for the $j$th local iteration of the global iteration $k$ on client $i$ of cluster $l$, respectively. Additionally, $F_k^{(l,i)}$ is the loss for client $i$ of cluster $l$ at the global iteration $k$. $\omega_k^{(l)}$ is the global shared network parameters on  IS $l$ at the beginning of the global iteration $k$, and $\beta$ is the learning rate for both the client local updates and the PS global updates. $FGN\_Server (\cdot)$ given in Algorithm~\ref{alg:fedgradnorm-server} performs the auxiliary loss $F_{\textrm{grad}}$ construction and minimization via gradient descent.     

\begin{algorithm}[]
    \caption{FGN\_Server$\left(\{\tilde{F}^{(l,i)}\}_{i=1}^N,\{g^{(l,i)}\}_{i=1}^N,\{p'^{(l,i)}\}_{i=1}^N \right)$}
    \label{alg:fedgradnorm-server}
\begin{algorithmic}[1]
    \State Construct the sparsified version of auxiliary loss function $F^{(l)}_{\textrm{grad}} \left(\{p^{(l,i)}\}_{i=1}^N \right)$ as given in eq. (\ref{F_grad_sparsified}) using $\{g^{(l,i)}\}_{i=1}^N$ and the loss ratios $\{\tilde{F}^{(l,i)}\}_{i=1}^N$.
    \State Update the loss weights by gradient descent $p^{(l,i)} \leftarrow p'^{(l,i)} - \alpha \nabla_{p^{(l,i)}} F^{(l)}_{\textrm{grad}}$, $\forall i \in [N]$.
\end{algorithmic}
\end{algorithm}

\section{Experimental Results}
In this section, we evaluate the performance of \emph{HOTA-FedGradNorm} compared to the naive equal weighting case by analyzing their loss functions.

\subsection{Dataset Specifications}
\emph{RadComDynamic} \cite{JagannathMTL} is used as a dataset for the simulations. This dataset contains 125,000 data points which have the following three attributes: (1) 6 modulation classes which are amdsb, amssb, ask, bpsk, fmcw, pulsed continous wave (PCW). (2) 8 signal types which are AM radio, short-range, radar-altimeter, air-ground-MTI, airborne-detection, airborne-range, ground-mapping. (3) Anomaly behaviour which is defined as having SNR lower than -4 dB since SNR can be a proxy for geo-location information. Low SNR is considered as a signal coming from an outsider.

\subsection{Hyperparameters and Model Specifications}
We consider $C$ (number of clusters) $=10$  and $N$ (number of clients inside each cluster) $=3$. Therefore, data points are divided among all 30 clients where each client owns a specific personalized data. We consider $\sigma_l^2$ in the channel gain vector $\mathcal{N}(0,\sigma_l^2)$ as 1 for all clusters. $H_{th}$ defined in  (\ref{power_vector}) is taken as $3.2 \times 10^{-2}$. For $F_{\textrm{grad}}$ construction given in Algorithm~\ref{alg:fedgradnorm-server}, $\gamma =0.6$ is used as a measure of task asymmetry. Additionally, $\alpha= 0.008$ is used as the learning rate for $F_{\textrm{grad}}$ optimization in Algorithm~\ref{alg:fedgradnorm-server} while the learning rate $\beta$ is taken as 0.0003 for network training in Algorithm~\ref{alg:hota-fedgradnorm}. We use ADAM optimizer for both network training and $F_{\textrm{grad}}$ optimization.

As a shared model, We use a 5-layer FC neural network as explained in Table~\ref{share_model_table}. Then, a simple linear layer is used as a personalized network for each client to map the output data to the corresponding class output.

\begin{table}[h!]
\begin{center}
\begin{sc}
\begin{tabular}{|c|}
\hline
Shared Network\\
\hline
FC(256, 512)\\
FC(512, 1024)\\
FC(1024, 2048)\\
FC(2048, 512)\\
FC(512, 256)\\
\hline
\end{tabular}
\end{sc}
\end{center}
\caption{Shared network model.}
\label{share_model_table}
\vspace*{-0.3cm}
\end{table}

\subsection{Results and Analysis}
\begin{figure}[]
 	\begin{center}
 	\subfigure[]{%
 	\includegraphics[width=0.49\linewidth]{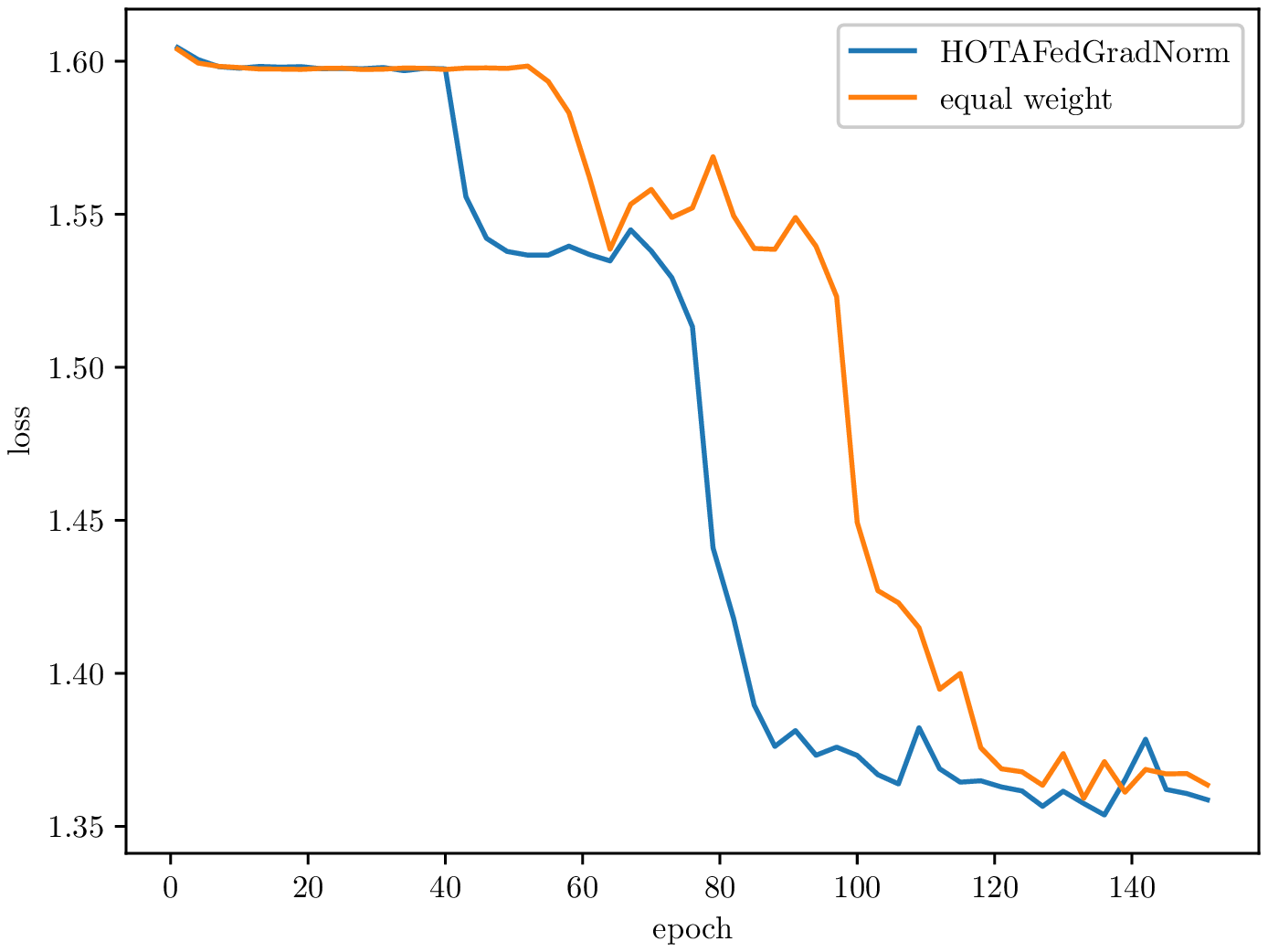}}
 	\subfigure[]{%
 	\includegraphics[width=0.49\linewidth]{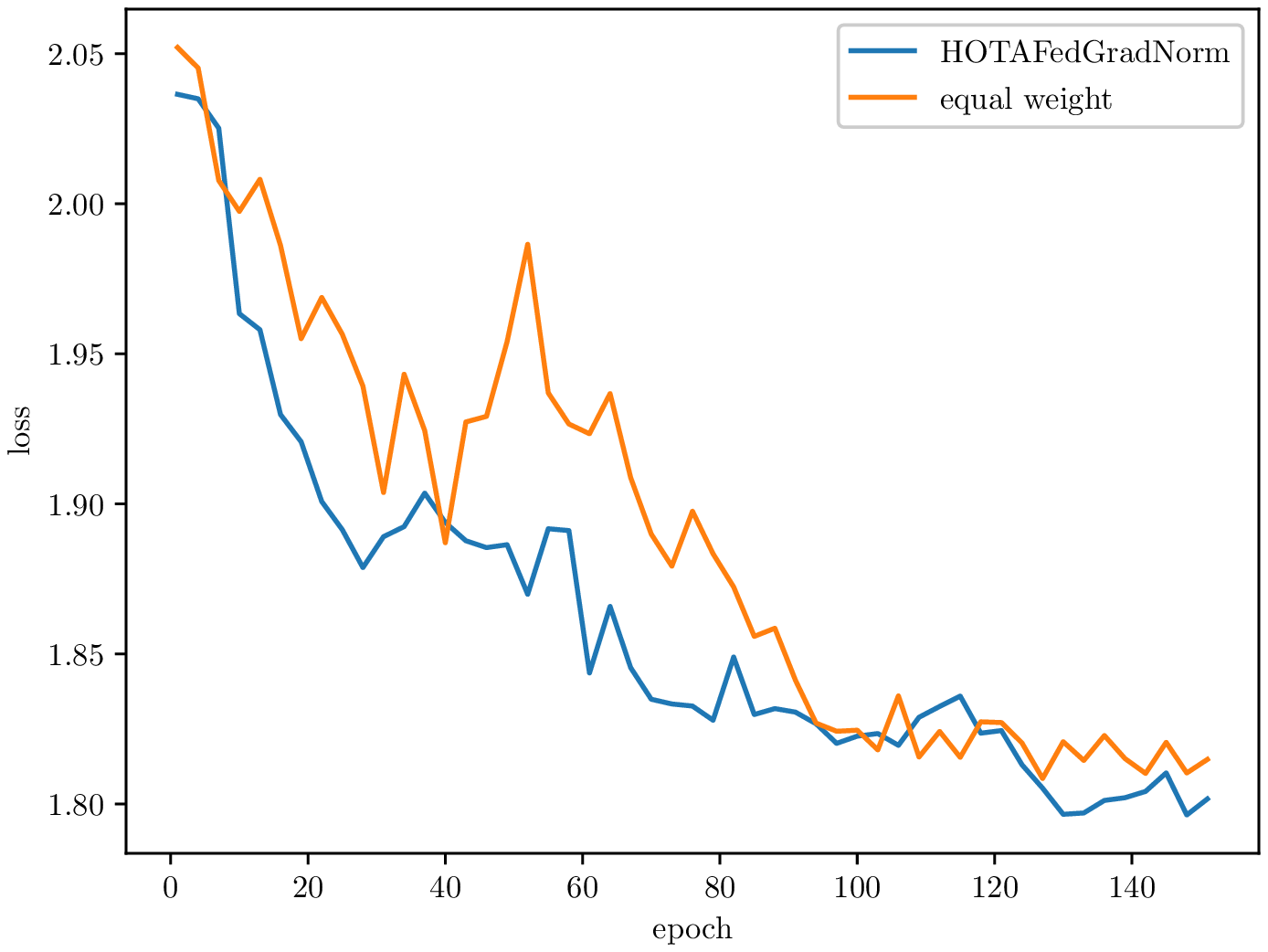}}\\ \vspace{-0.3cm}
 	\subfigure[]{%
 	\includegraphics[width=0.49\linewidth]{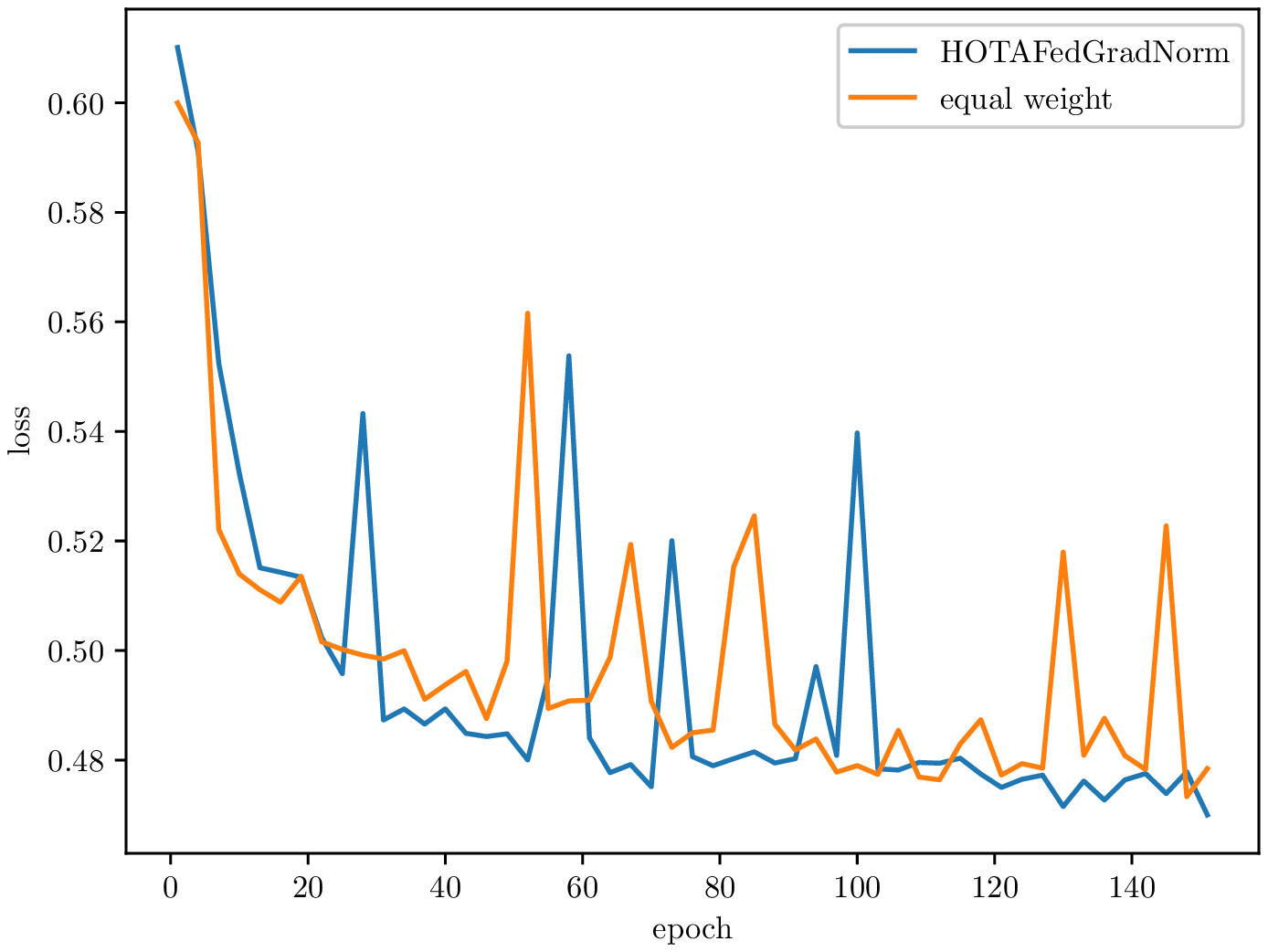}}
 	\subfigure[]{%
 	\includegraphics[width=0.49\linewidth]{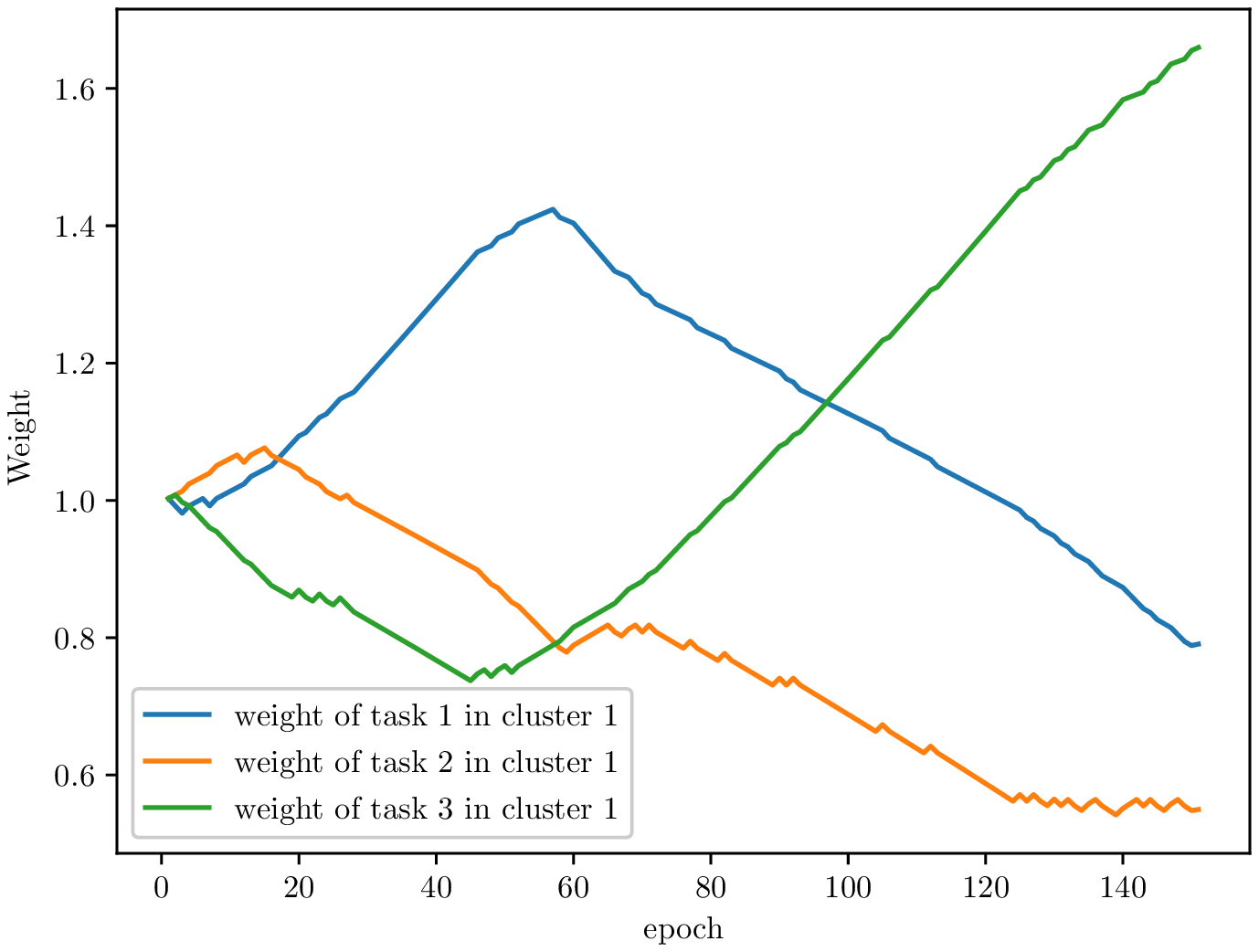}}
 	\end{center}
 	\caption{Comparison between task loss achieved via \emph{HOTA-FedGradNorm} and naive equal weighting case in RadComDynamic dataset for the first cluster (a) task 1 (modulation classification), (b) task 2 (signal classification), (c) task 3 (anomaly behavior), (d) task weights.}
 	\label{loss_HOTAFedgradnorm_wireless}
\end{figure}
Fig.~\ref{loss_HOTAFedgradnorm_wireless} depicts the task losses in the first cluster. The first task (modulation classification) has a lower change of the loss than the second and the third tasks' change of the loss at the beginning of the training. Therefore, we see an increase in the loss weight of the first task in all clusters. After epoch 65, the first task loss decreases significantly, thus, the corresponding loss weight is also decreased. 
Comparing the result with the result achieved in \cite{FedGradNorm}, we observe that considering the wireless MAC channel between the IS servers and the PS leads to slower training. However, as shown in Fig.~\ref{loss_HOTAFedgradnorm_wireless}, \emph{HOTA-FedGradNorm} yields a higher training speed compared to naive equal weighting strategy. 

\begin{figure}[]
 	\begin{center}
 	\subfigure[]{%
 	\includegraphics[width=0.49\linewidth]{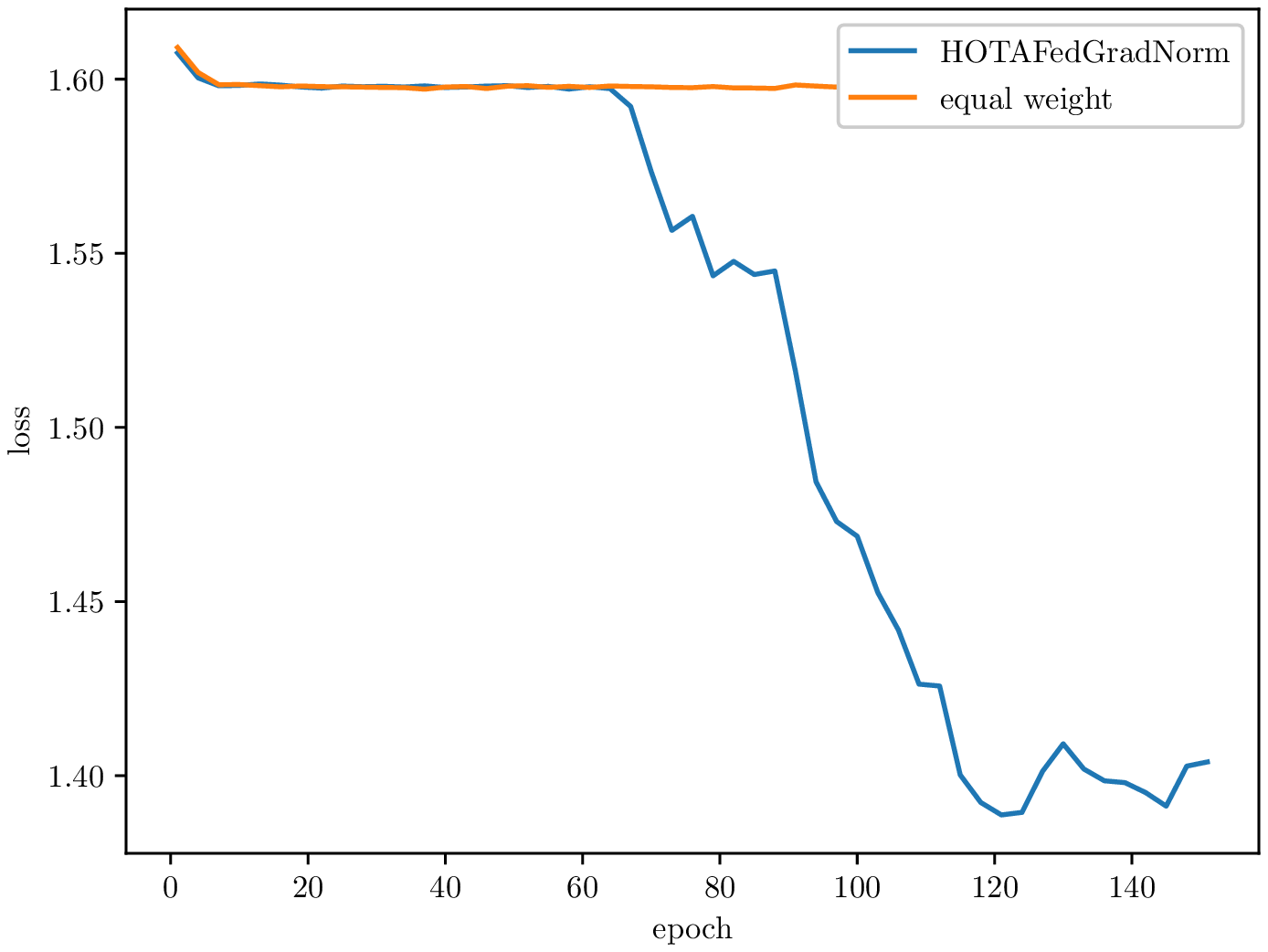}}
 	\subfigure[]{%
 	\includegraphics[width=0.49\linewidth]{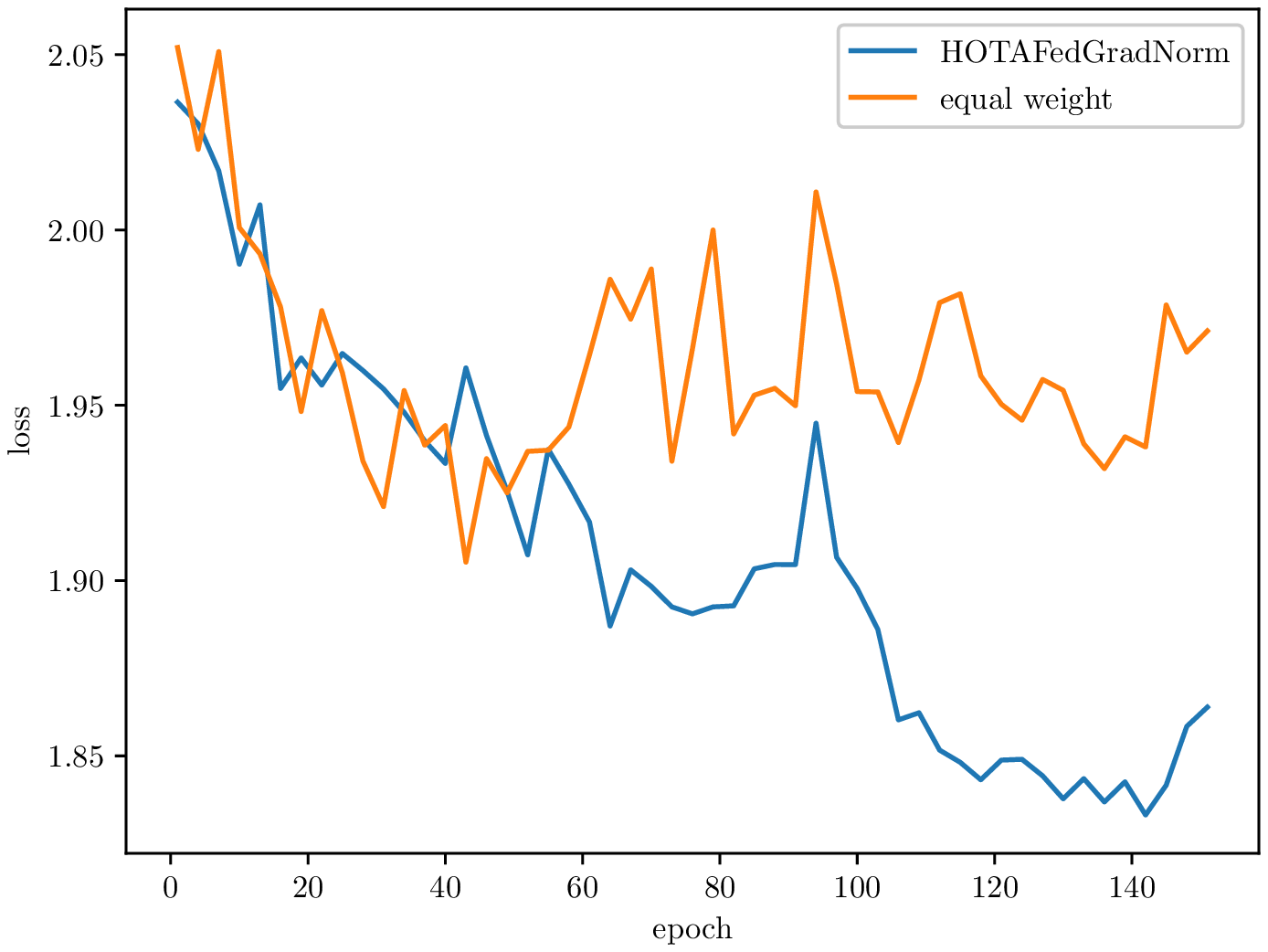}}\\ \vspace{-0.3cm}
 	\subfigure[]{%
 	\includegraphics[width=0.49\linewidth]{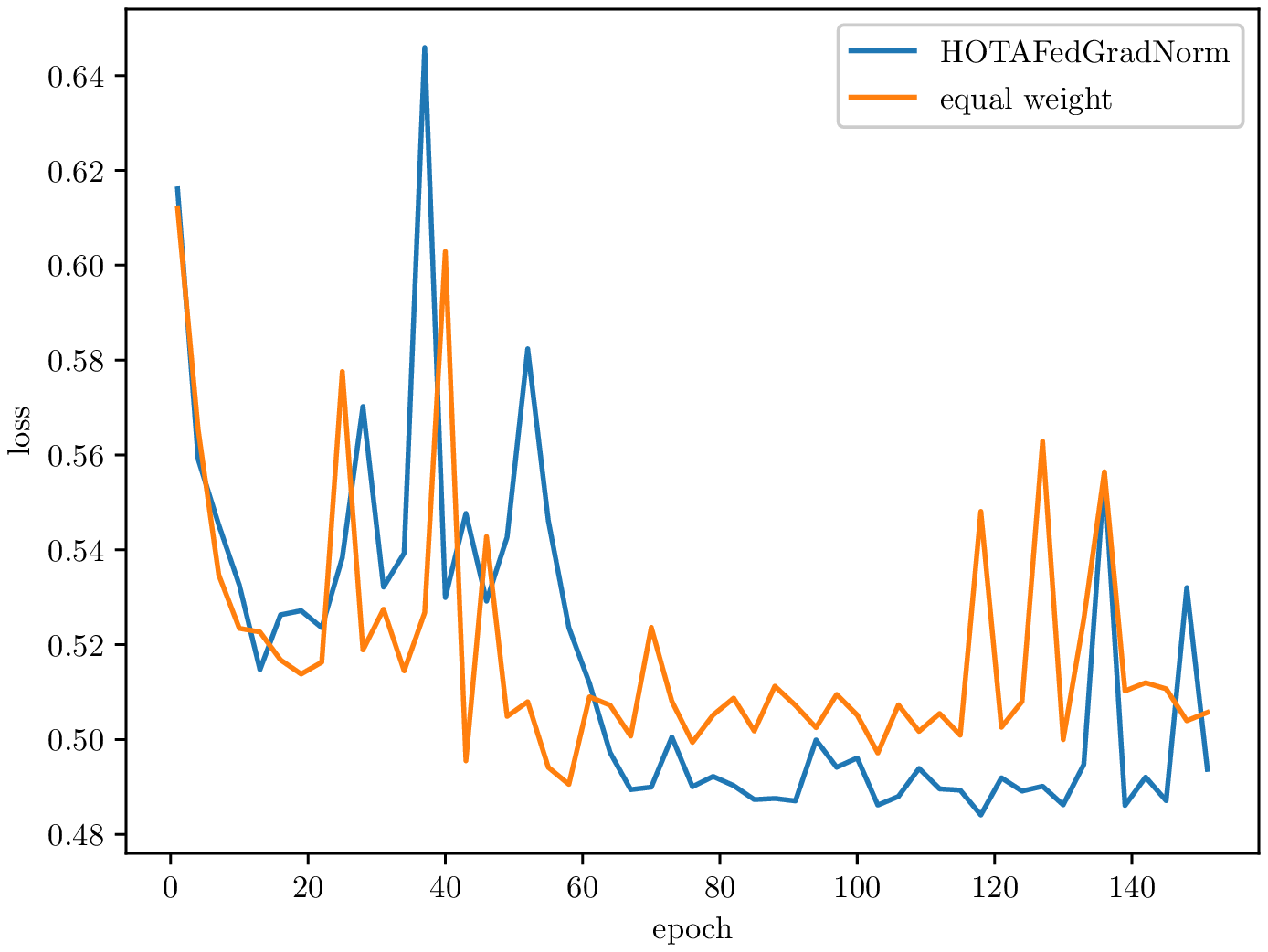}}
 	\subfigure[]{%
 	\includegraphics[width=0.49\linewidth]{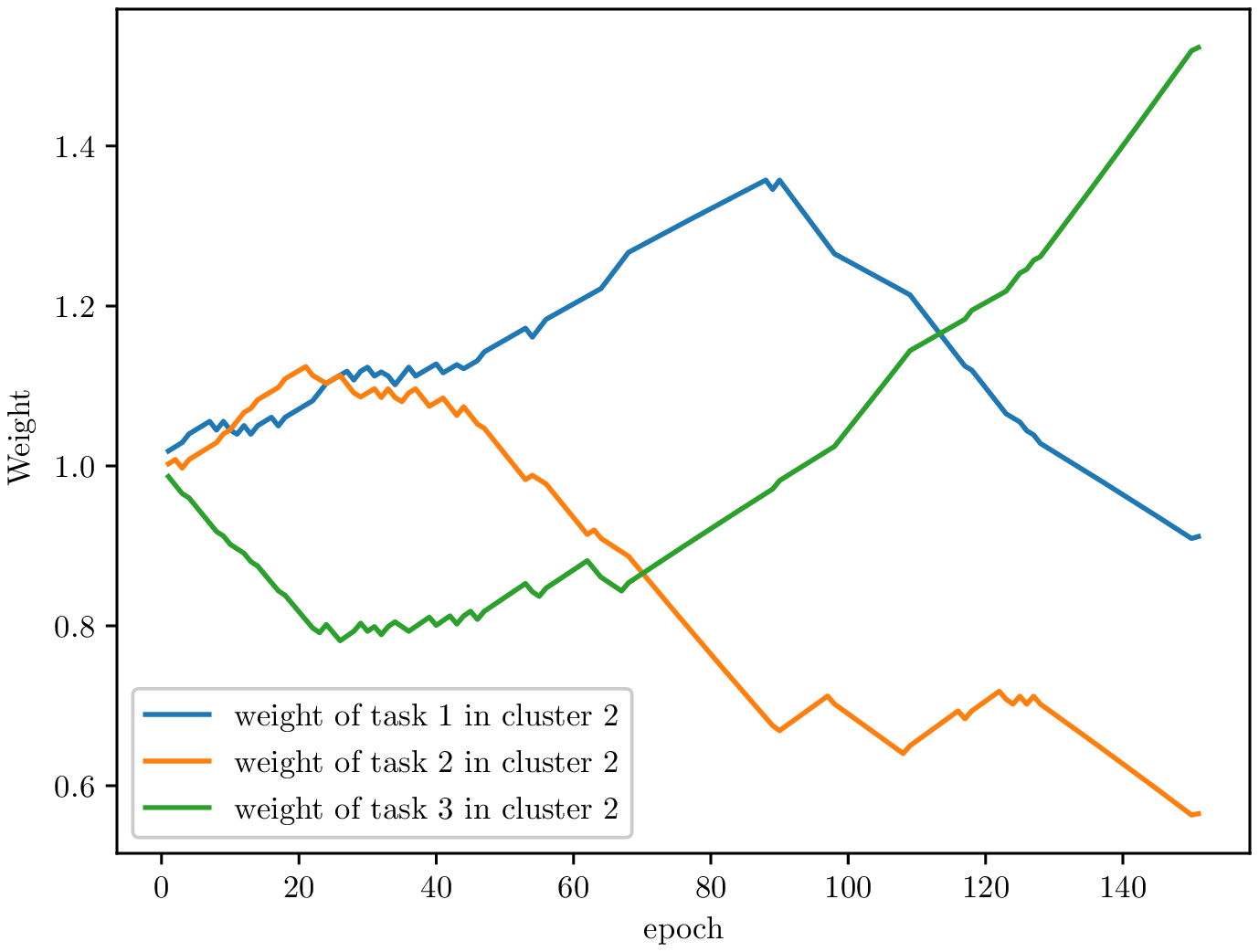}}
 	\end{center}
 	\caption{Comparison between task loss achieved via \emph{HOTA-FedGradNorm} and naive equal weighting case in RadComDynamic dataset for the second cluster where $\sigma_1^2 = 0.5$ and $\sigma_l^2 = 1$ $\forall l \geq 2$ (a) task 1 (modulation classification), (b) task 2 (signal classification), (c) task 3 (anomaly behavior), (d) task weights.}
 	\label{unbalance_loss_HOTAFedgradnorm_wireless}
\end{figure}

\begin{figure}[]
 	\begin{center}
 	\subfigure[]{%
 	\includegraphics[width=0.49\linewidth]{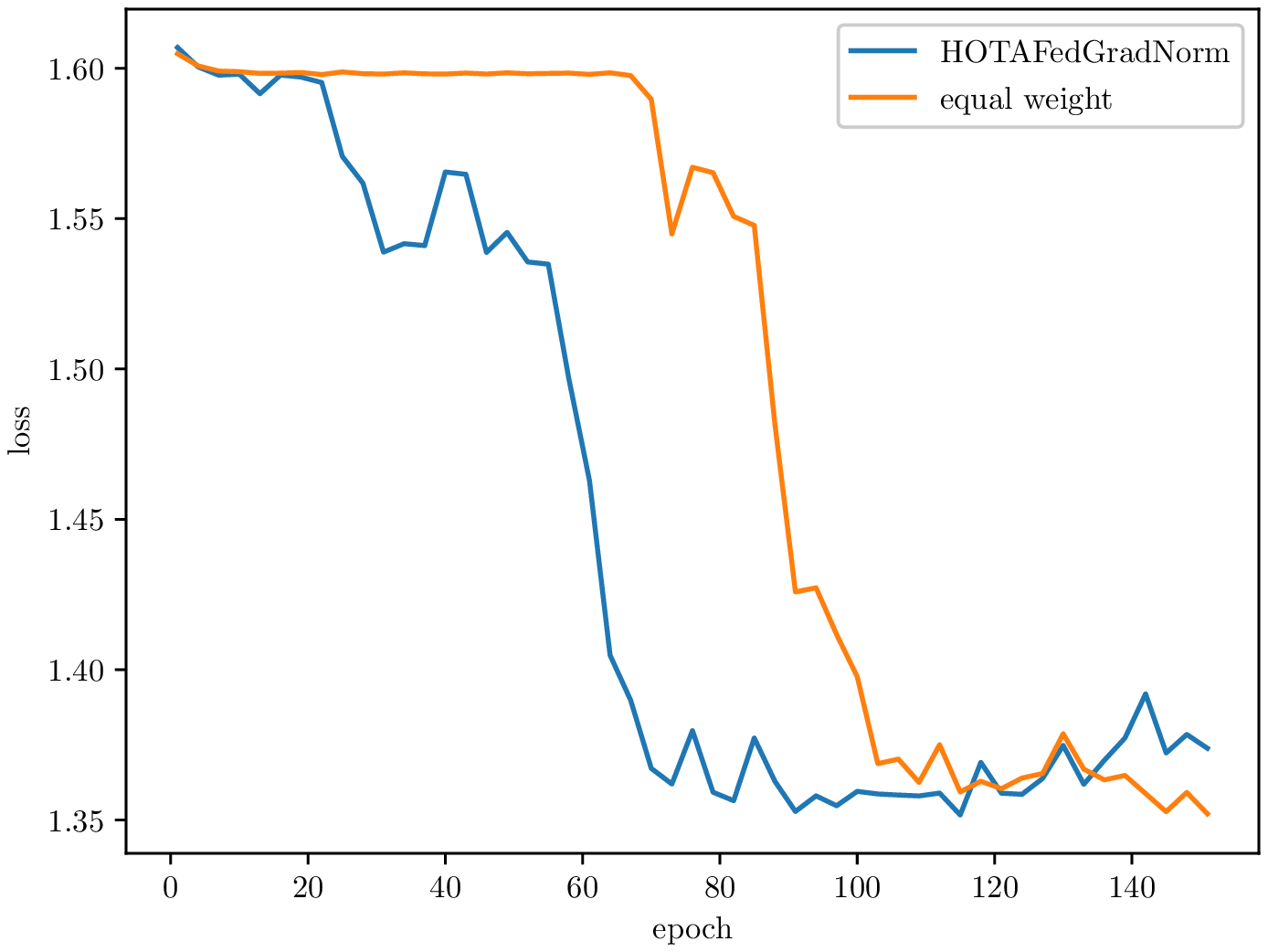}}
 	\subfigure[]{%
 	\includegraphics[width=0.49\linewidth]{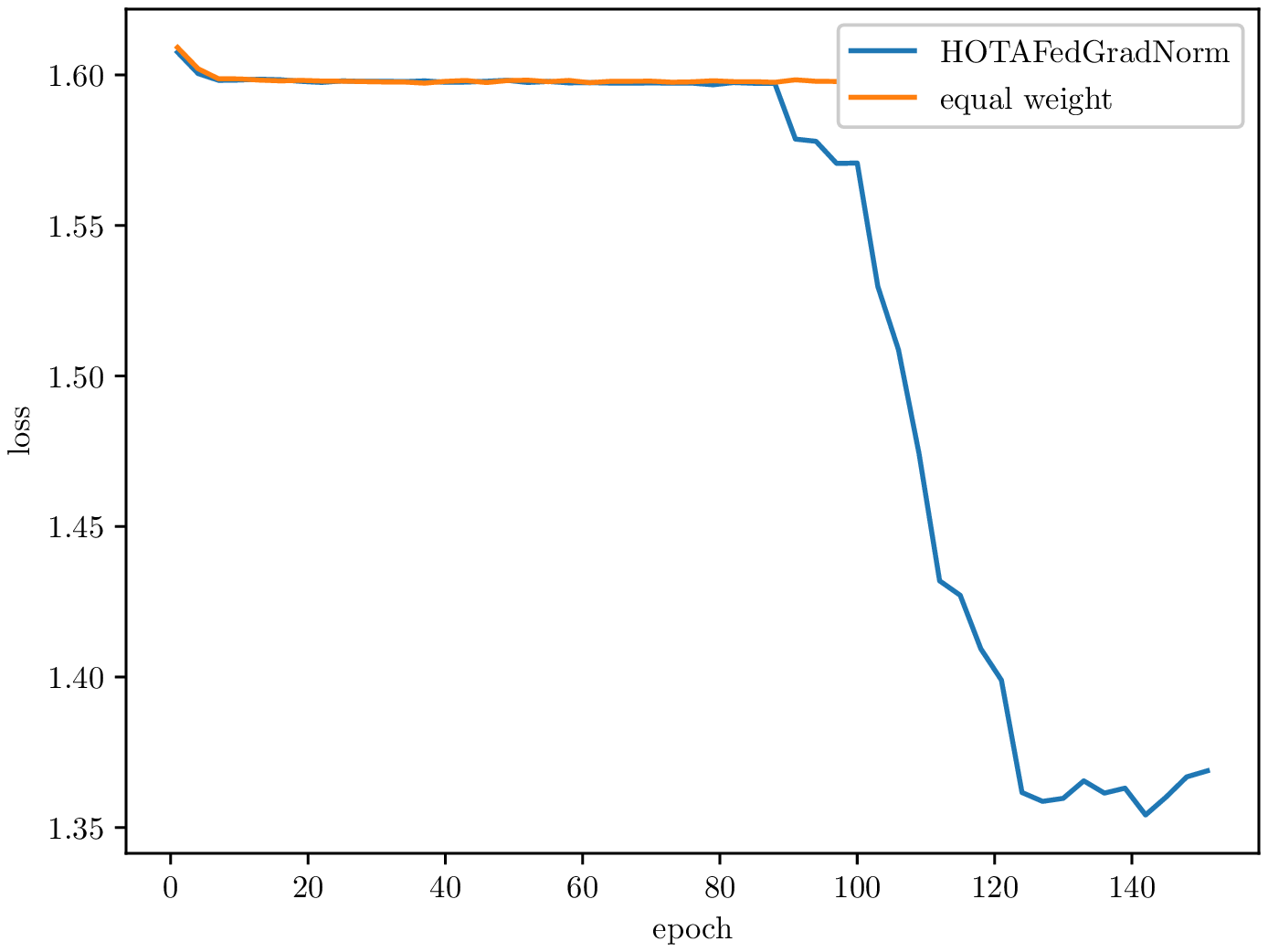}}\\ \vspace{-0.3cm}
 	\subfigure[]{%
 	\includegraphics[width=0.49\linewidth]{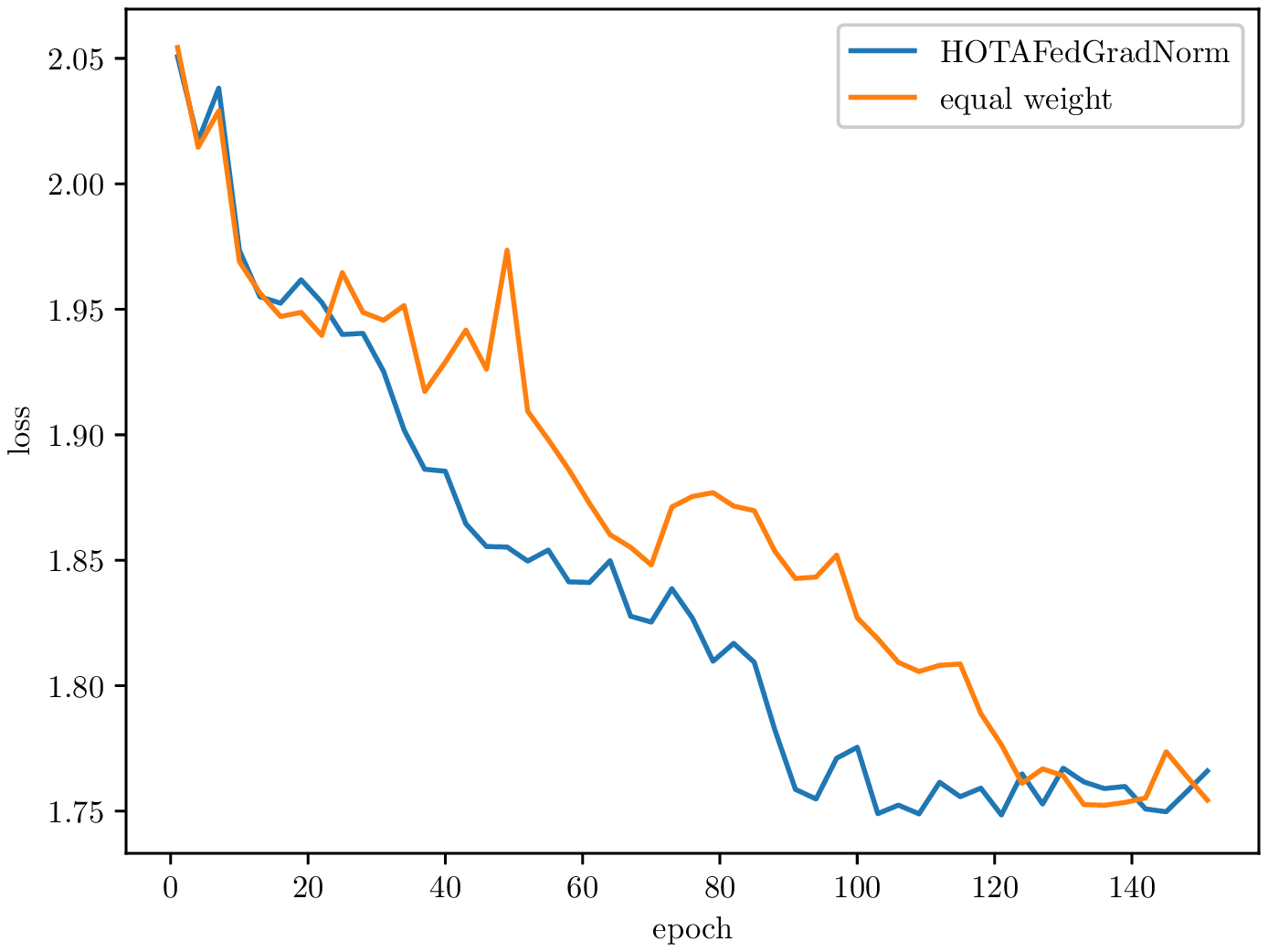}}
 	\subfigure[]{%
 	\includegraphics[width=0.49\linewidth]{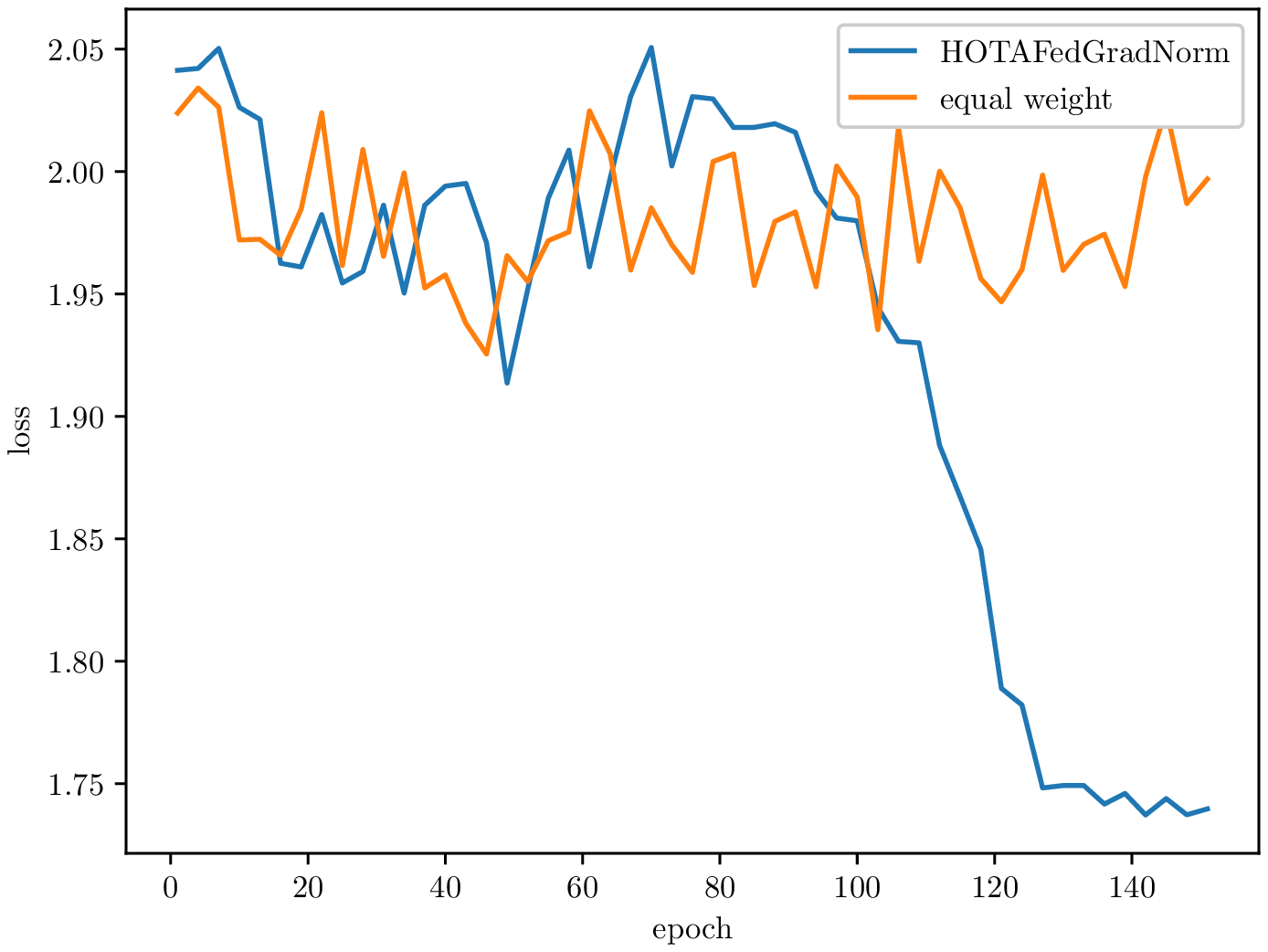}}
 	\end{center}
 	\caption{Comparison between task loss achieved via \emph{HOTA-FedGradNorm} and naive equal weighting case in RadComDynamic dataset where $\sigma_2^2 = 0.75$ and $\sigma_l^2 = 1$ for $\forall l \geq 3$ (a) task 1 (modulation classification) when $\sigma_1^2 = 2$, (b) task 1 (modulation classification) when $\sigma_1^2 = 0.25$, (c) task 2 (signal classification) when $\sigma_1^2 = 2$, (d) task 2 (signal classification) when $\sigma_1^2 = 0.25$,}
 	\label{comparison_HOTAFedgradnorm_wireless}
\end{figure}

To demonstrate the effectiveness of $F_{grad}$ to reduce the negative channel effects, we change the first cluster channel gain from $\sigma_1^2 = 1$ to $\sigma_1^2 = 0.5$ while keeping others as $\sigma_l^2=1$. Decreasing $\sigma_1^l$ value is equivalent to intensifying the sparsification of the updated gradient based on the definition of $H_{th}$.
Fig.~\ref{unbalance_loss_HOTAFedgradnorm_wireless} shows how even having a single bad channel in one cluster can deteriorate the entire learning performance if we do not incorporate \emph{FedGradNorm} into our system model. \emph{HOTA-FedGradNorm} modifies clients' weights based on the channel conditions, thereby, reducing the channel effects. Fig.~\ref{unbalance_loss_HOTAFedgradnorm_wireless} illustrates that both the first and second tasks are improved after epoch 85. Additionally, we compare the effects of channels for more diverse $\sigma$ values in Fig.~\ref{comparison_HOTAFedgradnorm_wireless}. We observe that \emph{HOTA-FedGradNorm} is both robust and faster to train under more challenging channel conditions.

\section{Conclusion}

In this paper, we proposed \emph{HOTA-FedGradNorm}, a hierarchical over-the-air personalized federated learning framework with a dynamic weighting strategy. Hierarchical federated learning and over-the-air aggregation approaches are integrated with distributed dynamic weighting to cope with the noisy channel in a power- and bandwidth-limited regime while sending updates from clients to the PS. We conducted experiments on the RadComDynamic dataset. We compared the experimental results between the dynamic weighting strategy and the naive equal weighting strategy for both mild and harsh channel conditions. The results demonstrated that \emph{HOTA-FedGradNorm} improves training speed in most tasks and provides robustness against the negative channel effects by considering the channel conditions during the dynamic weight selection process.

\bibliographystyle{unsrt}
\bibliography{reference2}
\end{document}